%% file: main.tex
\documentclass[10pt,twocolumn,letterpaper]{article}

\usepackage{cvpr}              %

\input{preamble}

\definecolor{cvprblue}{rgb}{0.21,0.49,0.74}
\usepackage[pagebackref,breaklinks,colorlinks,citecolor=cvprblue]{hyperref}

\def\paperID{10169} %
\def\confName{CVPR}
\def\confYear{2024}

\title{Constellation Dataset: Benchmarking High-Altitude Object Detection for an Urban Intersection}

\author{Mehmet Kerem Turkcan\\
Columbia University\\
New York, NY\\
{\tt\small mkt2126@columbia.edu}
\and
Chengbo Zang\\
Columbia University\\
New York, NY\\
{\tt\small cz2678@columbia.edu}
\and
Sanjeev Narasimhan\\
Columbia University\\
New York, NY\\
{\tt\small sn3007@columbia.edu}
\and
Gyung Hyun Je\\
Columbia University\\
New York, NY\\
{\tt\small gj2353@columbia.edu}
\and
Bo Yu\\
Columbia University\\
New York, NY\\
{\tt\small by2345@columbia.edu}
\and
Mahshid Ghasemi\\
Columbia University\\
New York, NY\\
{\tt\small mg4089@columbia.edu}
\and
Javad Ghaderi\\
Columbia University\\
New York, NY\\
{\tt\small jg3465@columbia.edu}
\and
Gil Zussman\\
Columbia University\\
New York, NY\\
{\tt\small gil.zussman@columbia.edu}
\and
Zoran Kostic\\
Columbia University\\
New York, NY\\
{\tt\small zk2172@columbia.edu}
}

\begin{document}
\maketitle
\input{sec/0_abstract}

\input{sec/1_intro}

\input{sec/2_related}
\input{sec/3_dataset}
\input{sec/4_experiments}

\input{sec/5_ethics_ack_conclusion}

{
    \small
    \bibliographystyle{ieeenat_fullname}
    \bibliography{main}
}

\end{document}

%% file: preamble.tex
\usepackage[dvipsnames]{xcolor}

%% file: sec/0_abstract.tex
\begin{figure*}[t!]
  \centering
  \begin{subfigure}{0.24\linewidth}
    \includegraphics[width=1.0\linewidth]{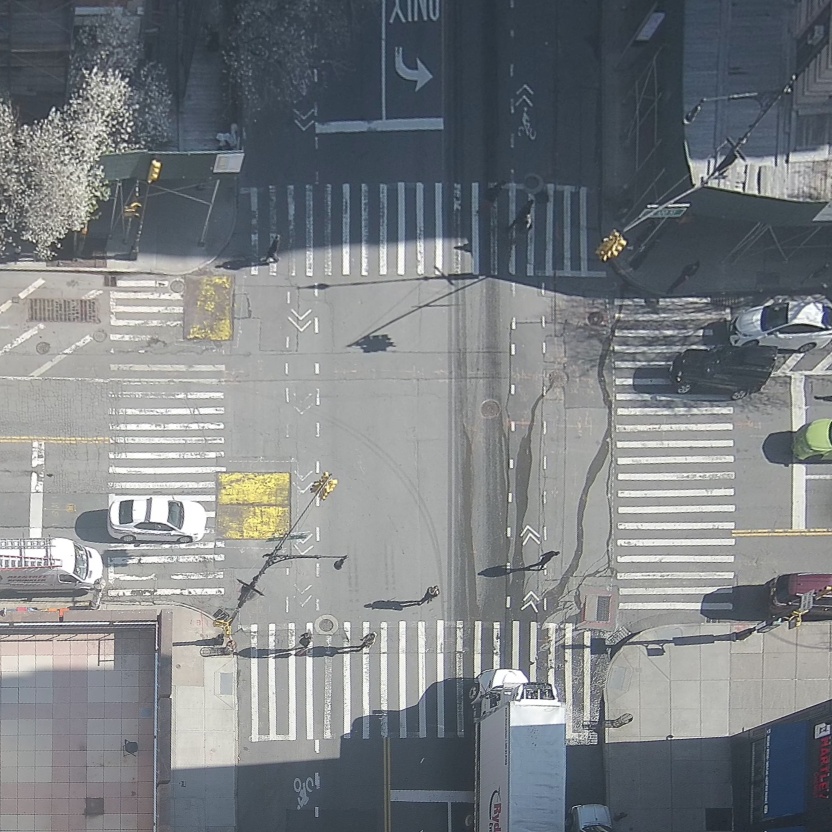}
    \caption{Dawn}
    \label{fig:short-c}
  \end{subfigure}
  \hfill
  \begin{subfigure}{0.24\linewidth}
    \includegraphics[width=1.0\linewidth]{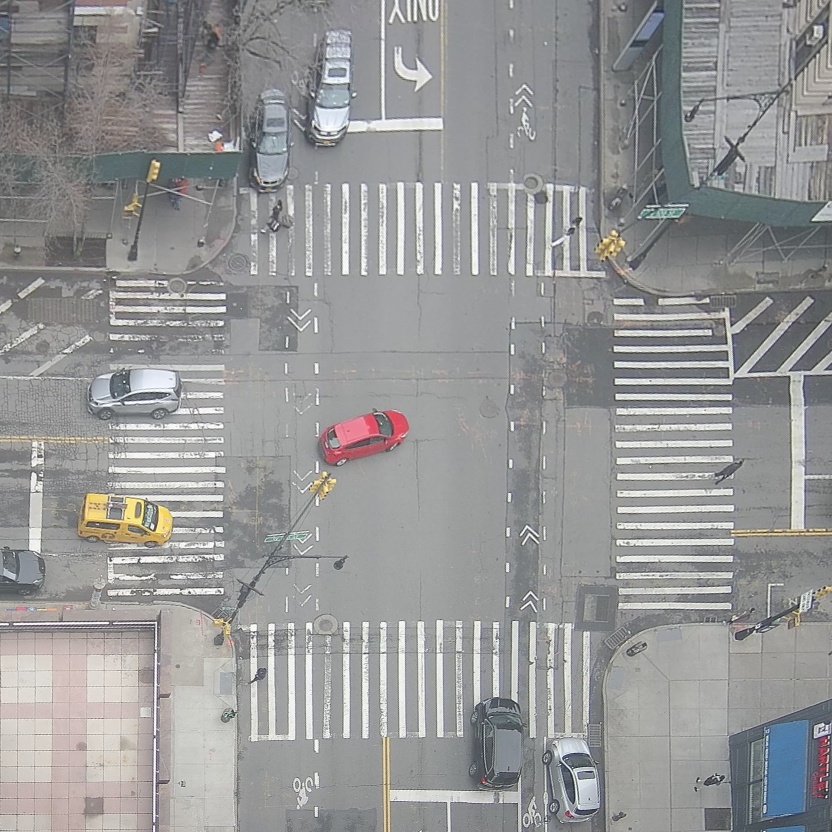}
    \caption{Daytime}
    \label{fig:short-a}
  \end{subfigure}
  \hfill
  \begin{subfigure}{0.24\linewidth}
    \includegraphics[width=1.0\linewidth]{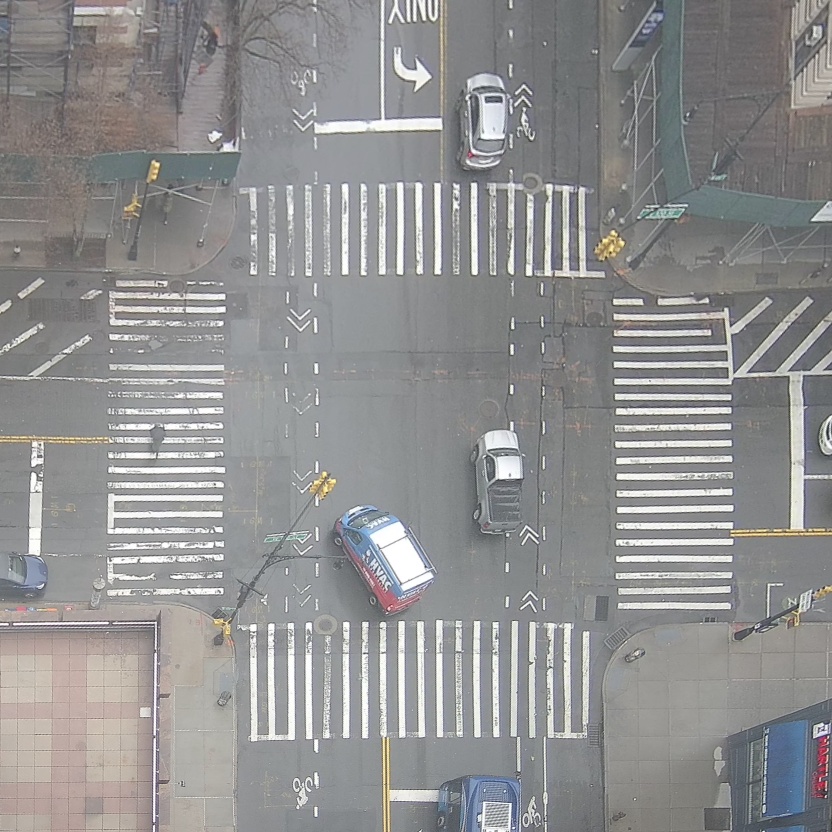}
    \caption{Rainy}
    \label{fig:short-b}
  \end{subfigure}
  \hfill
    \begin{subfigure}{0.24\linewidth}
    \includegraphics[width=1.0\linewidth]{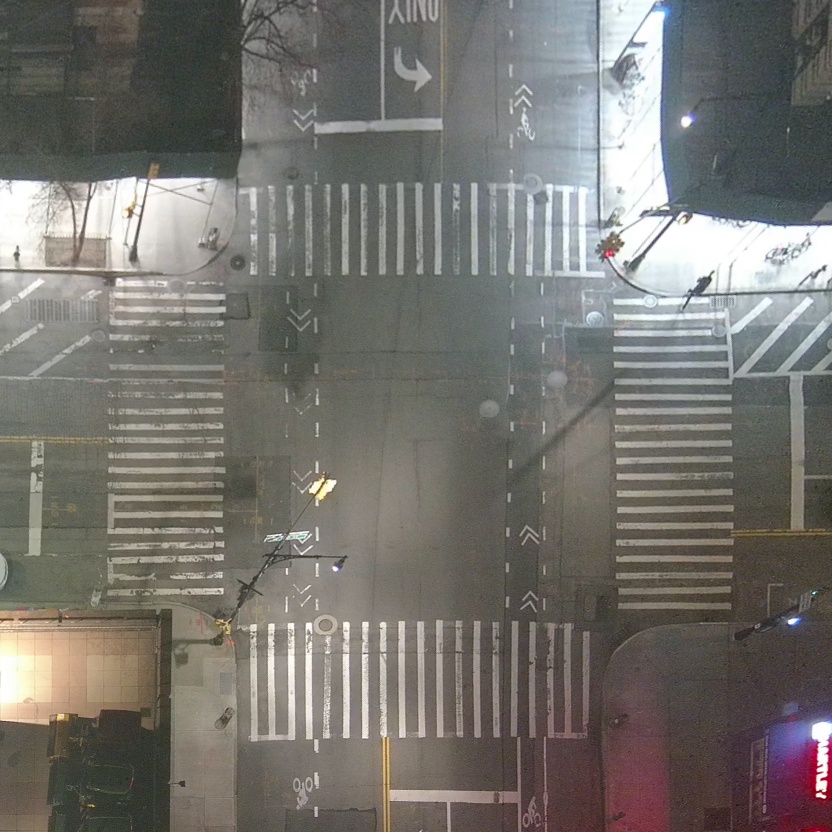}
    \caption{Night}
  \end{subfigure}\\
  \begin{subfigure}{0.24\linewidth}
    \includegraphics[width=1.0\linewidth]{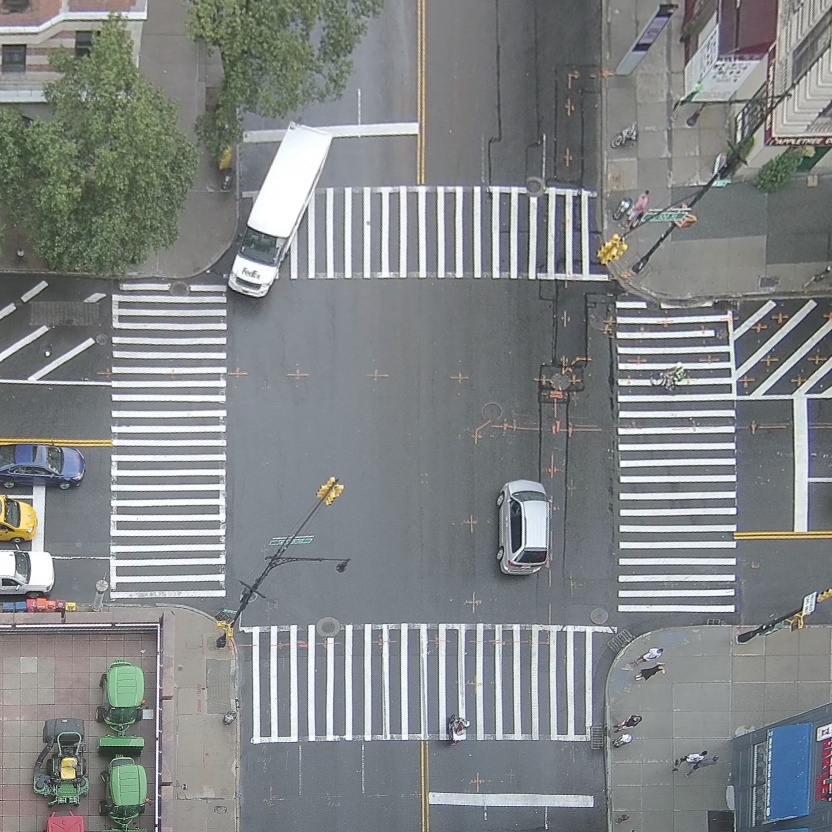}
    \caption{Old Pavement (2019)}
    \label{fig:short-a}
  \end{subfigure}
  \hfill
  \begin{subfigure}{0.24\linewidth}
    \includegraphics[width=1.0\linewidth]{figures/exs/rainy.jpg}
    \caption{Faded Pavement}
    \label{fig:short-b}
  \end{subfigure}
  \hfill
  \begin{subfigure}{0.24\linewidth}
    \includegraphics[width=1.0\linewidth]{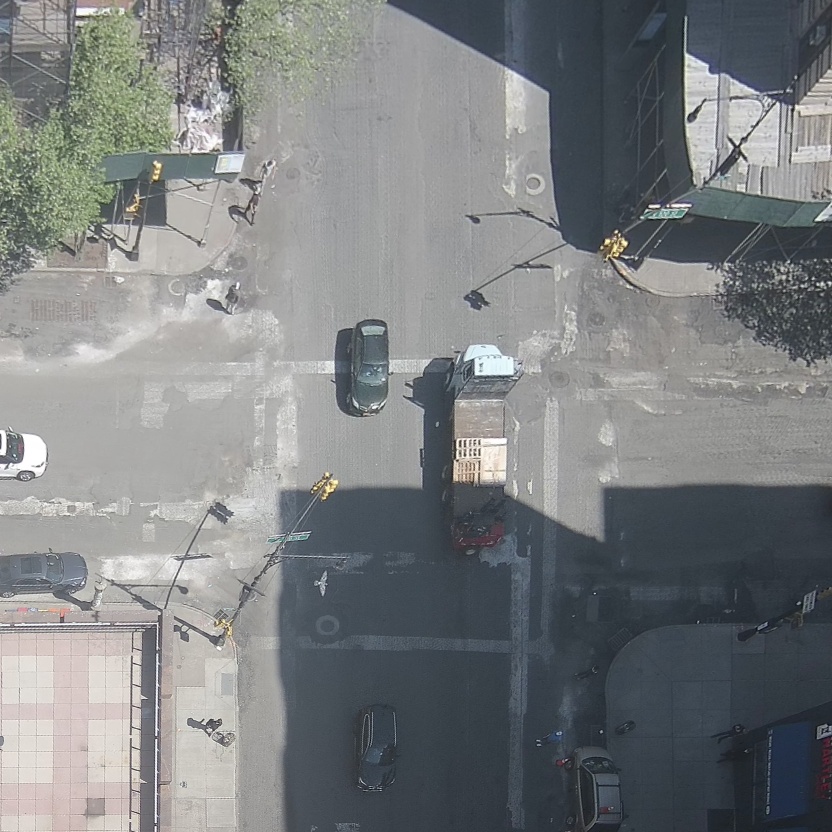}
    \caption{Unpaved}
    \label{fig:short-c}
  \end{subfigure}
  \hfill
    \begin{subfigure}{0.24\linewidth}
    \includegraphics[width=1.0\linewidth]{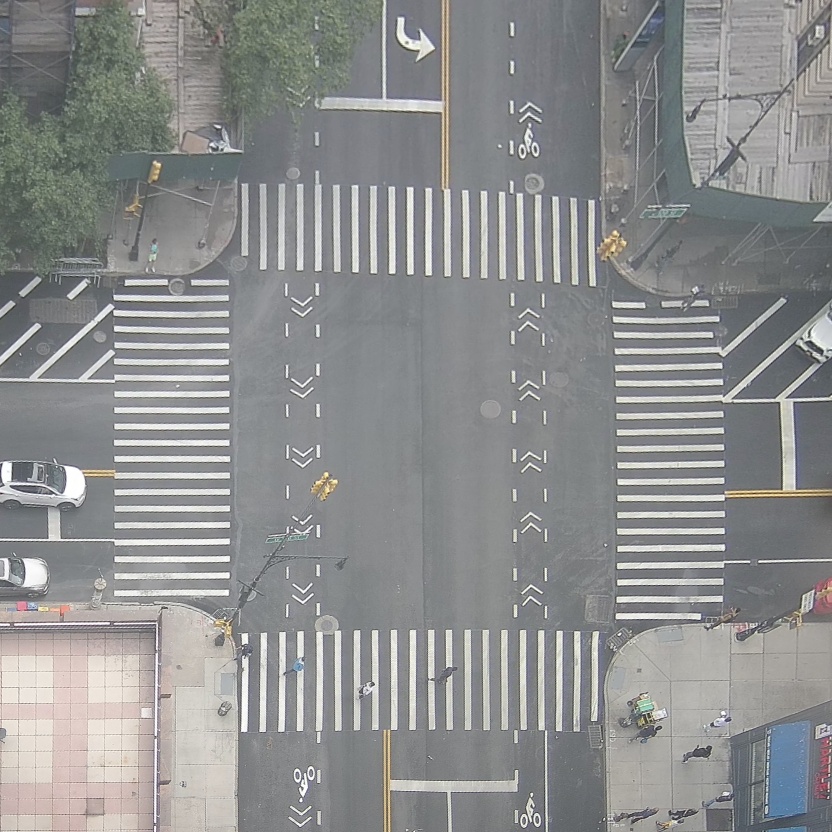}
    \caption{Repaved}
  \end{subfigure}
  \caption{Constellation contains different scenes, with changing time-of-day, weather conditions and background elements for the same camera. (a-d) Different weather and time-of-day conditions; (e-h) changes to the scene background.}
  \label{fig:constellation_overview}
\end{figure*}

\begin{abstract}

We introduce Constellation, a dataset of $13K$ images suitable for research on detection of objects in dense urban streetscapes observed from high-elevation cameras, collected for a variety of temporal conditions. 
The dataset addresses the need for curated data to explore problems in small object detection exemplified by the limited pixel footprint of pedestrians observed tens of meters from above. It enables the testing of object detection models for variations in lighting, building shadows, weather, and scene dynamics.  
We evaluate contemporary object detection architectures on the dataset, observing that state-of-the-art methods have lower performance in detecting small pedestrians compared to vehicles, corresponding to a 10\% difference in average precision (AP).
Using structurally similar datasets for pretraining the models results in an increase of 1.8\% mean AP (mAP). We further find that incorporating domain-specific data augmentations helps improve model performance. Using pseudo-labeled data, obtained from inference outcomes of 
the best-performing models, improves the performance of the models. 
Finally, comparing the models trained using the data collected in two different time 
intervals, we find a performance drift in models due to the changes in intersection conditions over time.
The best-performing model achieves a pedestrian AP of 92.0\% with 11.5 ms inference time on NVIDIA A100 GPUs, and an mAP of 95.4\%. 
\end{abstract}

%% file: sec/1_intro.tex
\section{Introduction}
\label{sec:intro}

Urban landscapes are transitioning towards ``smart cities" with the underlying aim of catering to societal welfare~\cite{sanchez2014smartsantander,belli2020iot}. 
Considering deployment challenges and low-latency solution requirements, traffic intersections stand out as optimal locations for embedding intelligence nodes in smart cities. A city's operations can be enhanced through the interlinking of these nodes situated at adjacent intersections.

Dense urban locales are particularly challenging for autonomous vehicles because of occlusions, high density of car and pedestrian traffic, multitude of vehicles navigating in diverse trajectories at varied velocities, physical obstructions, and the unpredictable movements of pedestrians. The layout of buildings and their electrical power and communications infrastructure present opportunities to install 
sensors on city buildings, use V2X and related technologies and thereby upgrade autonomous vehicles to cloud-connected vehicles~\cite{chen2017vehicle,galgano2021connected}. The ultimate goals are a dramatic increase in pedestrian safety and improvements in traffic flow.

Utilizing cameras to monitor and track pedestrian and vehicular movement is an effective solution for traffic surveillance and increasing safety at smart intersections. A great majority of cameras in cities are positioned no higher than the first floor, which presents complications for object detection associated with occlusions and the intricacies of top-down view transformations. In major metropolises, cameras can be positioned at higher altitudes, which facilitates easier generation of bird's-eye perspective images through low-latency perspective transformations.

The overwhelming majority of the available datasets for object detection in urban settings are collected using low-altitude cameras. This is convenient from the perspective of cameras' installations but frequently results in visual occlusions of objects.

In this study, we introduce Constellation, a dataset containing $13,314$ human-annotated images with bounding box annotations for pedestrians and vehicles captured at an intersection in New York City using high-elevation cameras on the COSMOS testbed \cite{raychaudhuri2020challenge}. This dataset offers the ability to study pedestrian and vehicle detection in a real-world urban environment, enabling the development of robust algorithms that can handle the challenges posed by diverse weather conditions, lighting variations, and complex scenes. Constellation is the first dataset to feature an actual deployment in an urban metropolis, spanning multiple years, seasons, weather conditions, and times of day.

Using the Constellation dataset, we compared contemporary object detection models and found that single-stage YOLO models~\cite{terven2023comprehensive} are the best-performing in terms of inference time and precision.
Transformer-based RT-DETR models and Faster R-CNN-based CFINet obtained poorer results~\cite{lv2023detrs,yuan2023small}. We then investigated
pretraining based on other similar datasets and custom augmentations to improve model performance. We found that using pretraining datasets improves the model performance on our Constellation dataset by 1.8\%\thinspace{mAP}, providing a visible increase in pedestrian detection performance. We used 
the best-performing VisDrone-pretrained YOLOv8x model to automatically label data from the intersection
and used this newly acquired pseudo-labeled data as a pretraining dataset, showing that this approach is competitive against using external datasets for pretraining.
We release the codebase and baseline models trained with Constellation.

%% file: sec/2_related.tex
\section{Related Works}
\label{sec:related_works}
\textbf{Pedestrian and Vehicle Detection} A wealth of datasets from low-altitude on-vehicle or stationary cameras for vehicle and pedestrian detection have been released~\cite{geiger2013vision,oh2011large,sun2020scalability,personpath22}. Largely, these datasets do not feature scenes of urban metropolises and focus either on vehicles or pedestrians. Many of the publicly available traffic datasets specifically feature footage captured from street-level elevations. There exist datasets for drone-based high-altitude object detection, which largely focus on campus areas or highways rather than urban streetscapes, and some of which lack annotated bounding boxes needed for object detection
\cite{kovvali2007video,xia2018dota,robicquet2016learning,krajewski2018highd,yang2019top,krajewski2020round}. Images from low-altitude cameras are friendly for object detection, segmentation, and tracking applications since vehicles and pedestrians all occupy a large enough number of pixels (greater than $64$x$128$). Sizeable pixel areas make it easy for convolutional networks to capture high-quality feature maps and are present heavily in standard benchmark datasets like COCO~\cite{lin2014microsoft}. 

\textbf{Real-Time Object Detection} For real-time object detection, single-stage object detectors based on SSD or YOLO have gained popularity in recent years~\cite{liu2016ssd,redmon2016you,redmon2017yolo9000,redmon2018yolov3,2020arXiv200410934B,wang2022yolov7,xu2022pp,yolov8_ultralytics}. Unlike two-stage detectors that feature region proposal networks which propose potential bounding boxes, single-stage detection models generate a fixed number of bounding box proposals based on integrated low-dimensional feature maps. Recently proposed transformer-based architectures like the detection transformer (DETR) lack the real-time performance required for deployment~\cite{carion2020end}. Recently, Real-Time Detection Transformer (RT-DETR), which replaces the encoder in DETR with a hybrid encoder, has been proposed as an alternative to YOLO architectures~\cite{lv2023detrs}.

\textbf{Small Object Detection} Object detection models typically struggle with small object detection. Tiny pedestrians in a  bird's eye camera view are represented by inadequately precise feature maps. This results in poor detection and tracking accuracy.
Many approaches have been proposed to improve small object detection, especially in the context of aerial object detection captured by VisDrone and DOTA datasets \cite{zhu2018visdrone,xia2018dota,wang2021tiny,xu2021dot}. We note that aerial object detection datasets contain significantly different views than the static high-altitude infrastructure-colocated camera placements which we consider in this study.

For YOLO models, adding extra levels of the Feature Pyramid Network (FPN) has been suggested to improve the size of the feature maps, and thus the capability to increase small object detection accuracy~\cite{lai2023stc}. Recently, RTMDet proposed modifying the YOLO architecture by integrating large-kernel depth-wise convolutions and soft labels for aerial data \cite{lyu2022rtmdet}. PP-YOLOE models use several architectural choices to improve upon YOLO \cite{xu2022pp}. LSKNet proposes large and selective kernel modules to give models the contextual information needed \cite{li2023large}. Spatial Transform Decoupling improves upon ViT-based object detectors by estimating spatial transform parameters through separate network branches \cite{yu2023spatial}. Changing the resolution of the input through rescaling, slicing, super-resolution, or changing the model resolution, has been explored extensively for improving the performance of object detectors~\cite{shermeyer2019effects,chen2020survey,akyon2022slicing}. 

In this paper, we bring forward a new dataset aimed at high-altitude object detection in urban environments. We seek to motivate models that work in this real-world scenario that differs from existing datasets. Aerial drone-based imagery fundamentally differs from our use case due to the difference in capture height and stationarity of the camera; in many cities, urban laws restrict the collection of drone imagery, making it unsuitable for real-world deployment. High-altitude infrastructure-colocated object detection uses significantly different angles than aerial high-altitude views and requires building models robust to changes in conditions. 
The data split of the Constellation dataset accounts for a variety of environmental and time-of-day conditions, across several years of both training and testing, making it fundamentally different from other datasets. We evaluate Constellation on state-of-the-art models for aerial object detection.

\begin{figure*}[h!]
  \centering
  \begin{subfigure}{0.49\linewidth}
    \includegraphics[width=0.5\linewidth]{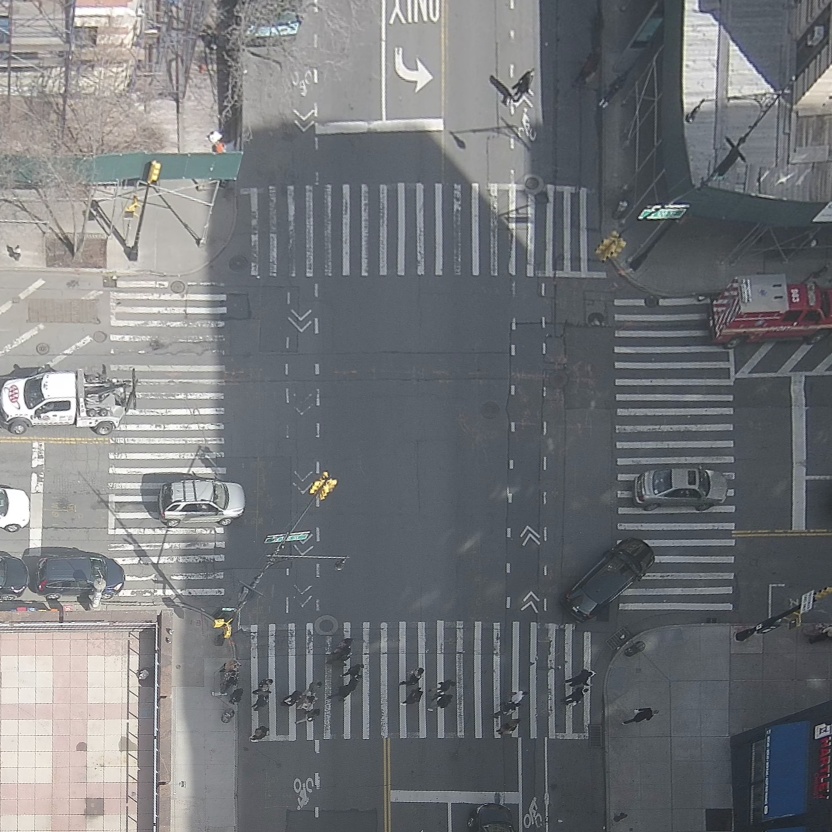}~
    \includegraphics[width=0.5\linewidth]{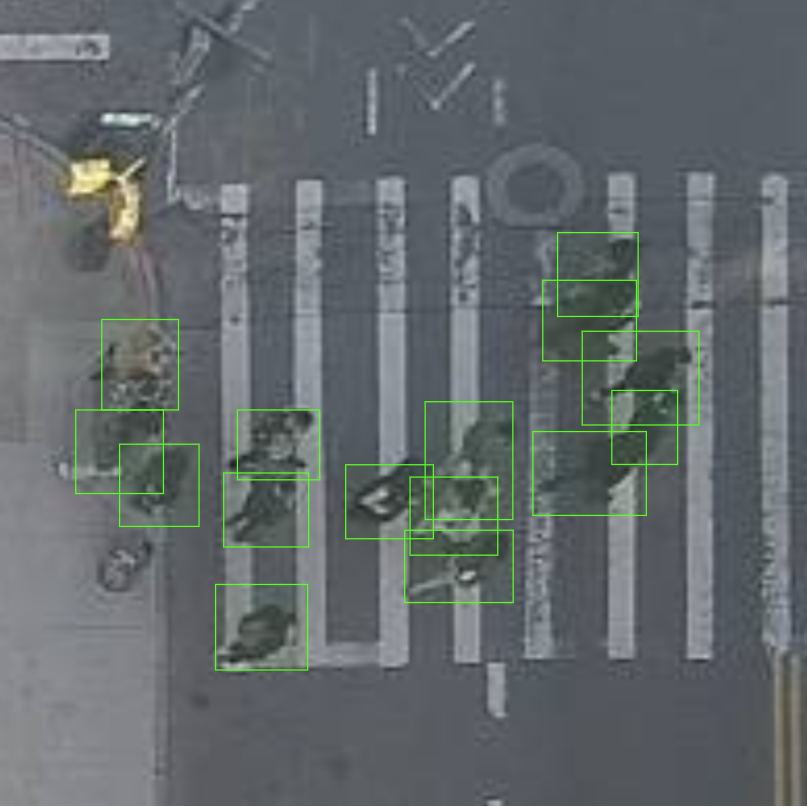}
    \caption{}
    \label{fig:short-b}
  \end{subfigure}
  \hfill
  \begin{subfigure}{0.49\linewidth}
    \includegraphics[width=0.5\linewidth]{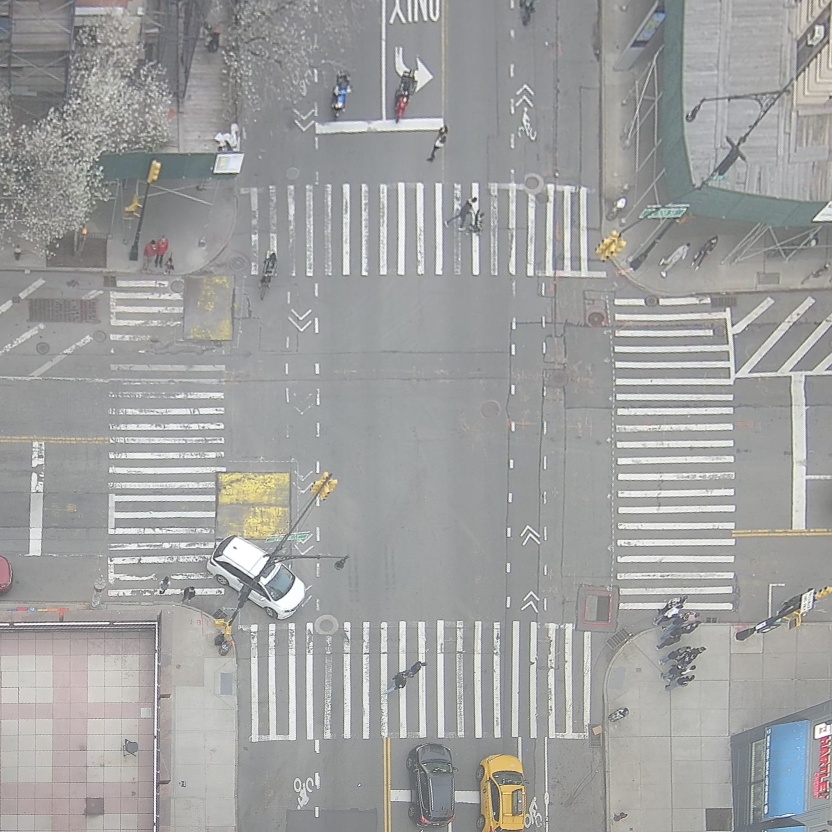}~
    \includegraphics[width=0.5\linewidth]{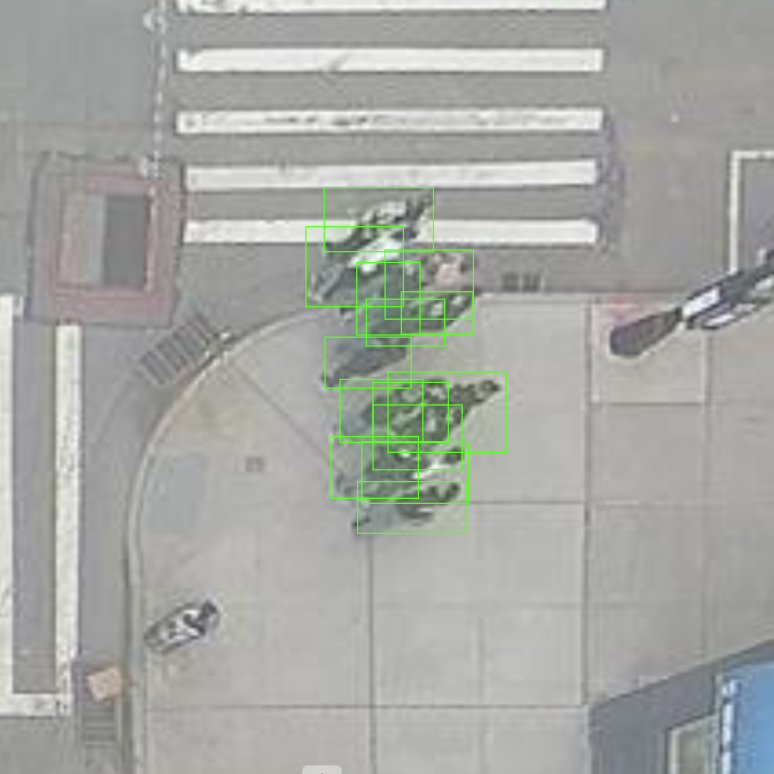}
    \caption{}
  \end{subfigure}
  \caption{Examples of two complex crowded frames from Constellation.}
  \label{fig:high_overlap}
\end{figure*}

\begin{figure*}[h!]
  \centering
  \begin{subfigure}{0.24\linewidth}
    \includegraphics[width=1.0\linewidth]{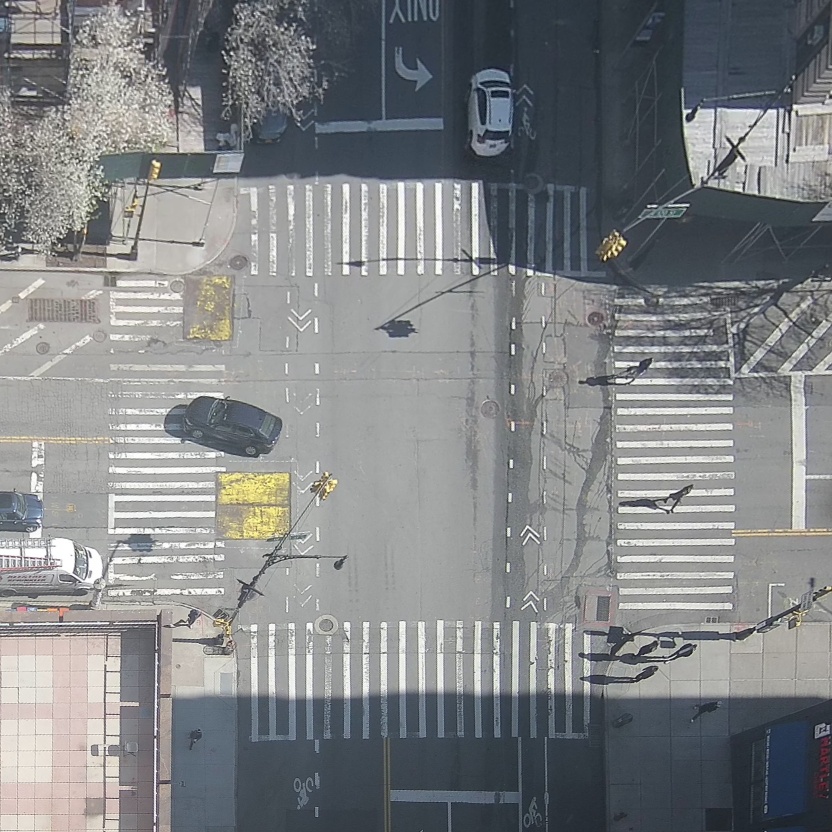}
    \caption{Constellation}
    \label{fig:short-a}
  \end{subfigure}
  \hfill
  \begin{subfigure}{0.24\linewidth}
    \includegraphics[width=1.0\linewidth]{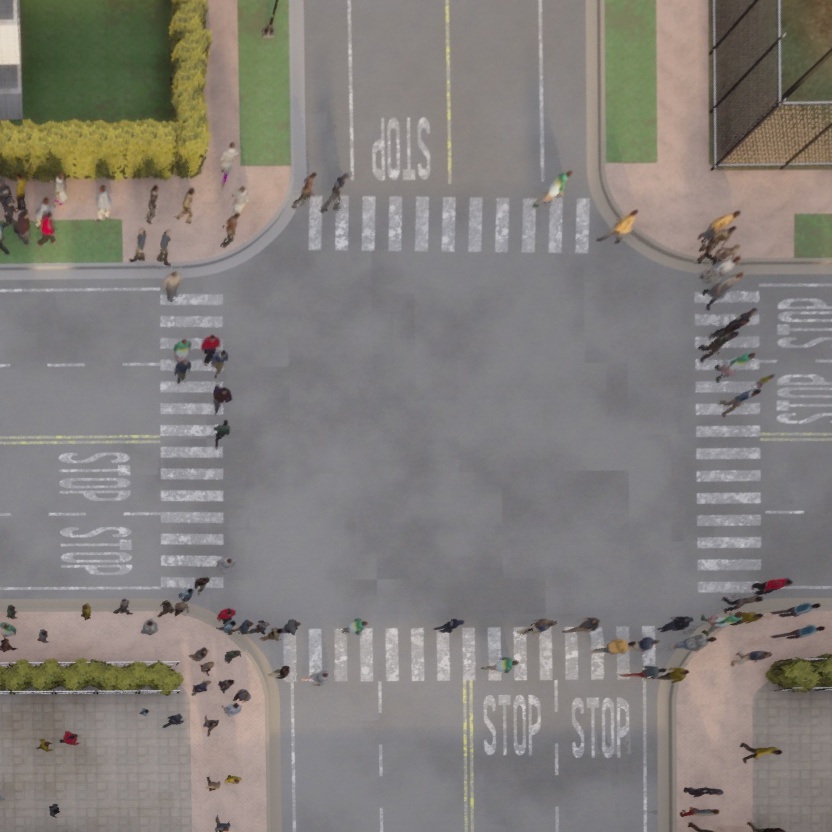}
    \caption{CARLA}
    \label{fig:short-b}
  \end{subfigure}
  \hfill
  \begin{subfigure}{0.24\linewidth}
    \includegraphics[width=1.0\linewidth]{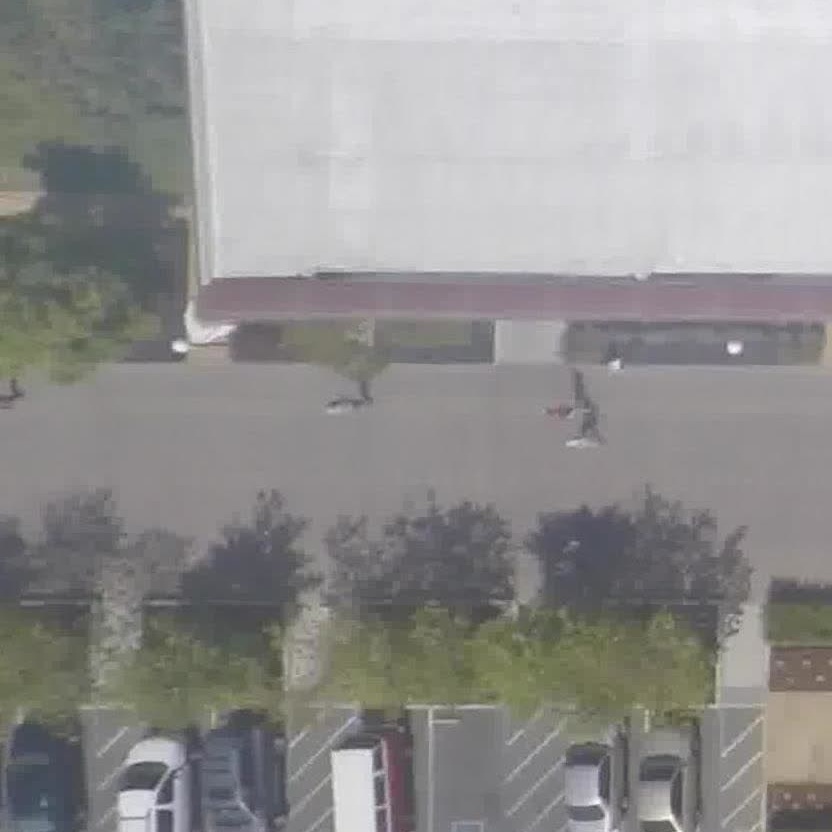}
    \caption{Stanford Drone Dataset}
    \label{fig:short-c}
  \end{subfigure}
    \begin{subfigure}{0.24\linewidth}
    \includegraphics[width=1.0\linewidth]{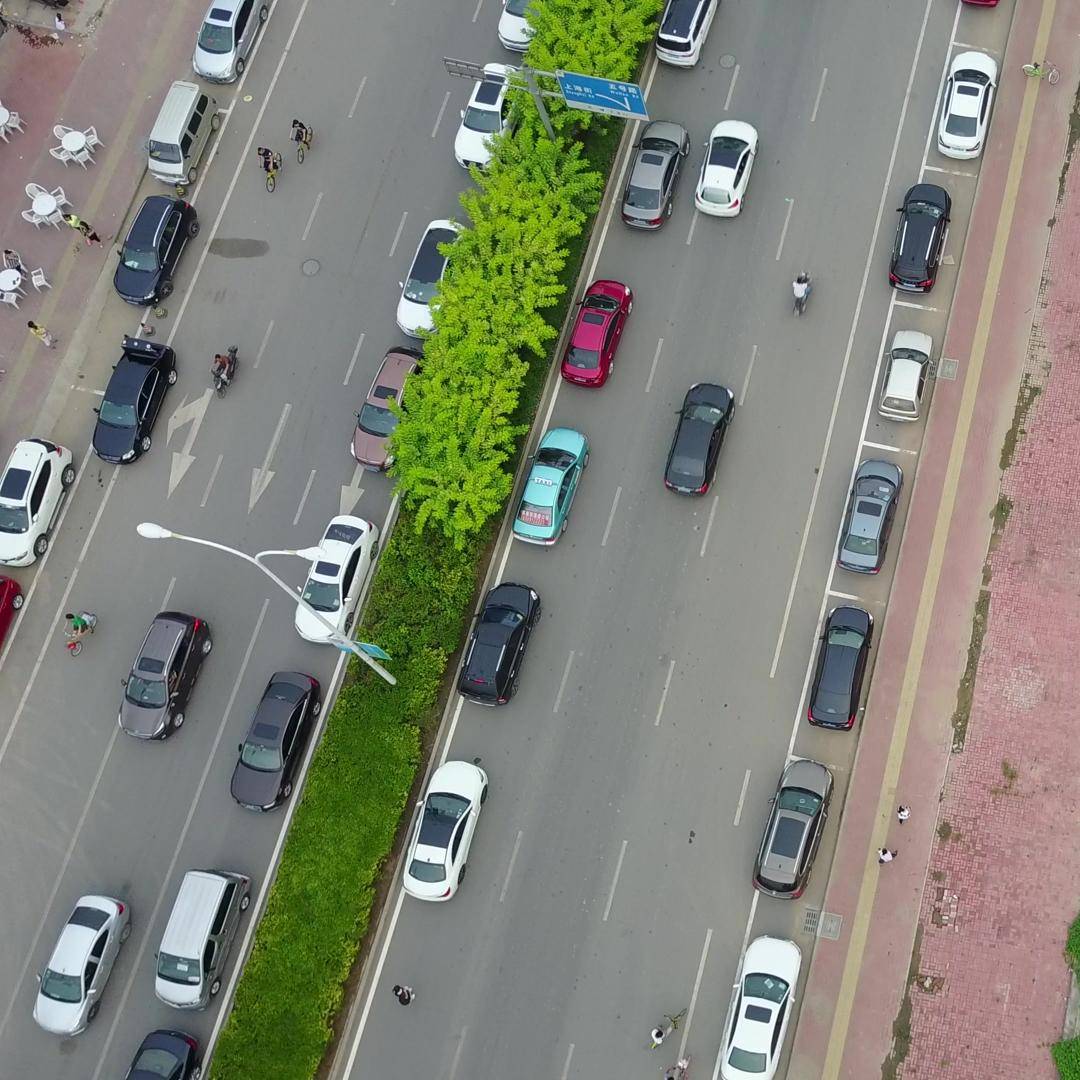}
    \caption{VisDrone}
    \label{fig:short-d}
  \end{subfigure}
  \caption{Different data sources used for experiments: (a) Constellation; (b) CARLA; (c) Stanford Drone Dataset; (d) VisDrone.}

  \label{fig:different_datasets}
\end{figure*}
\begin{figure*}[h!]
  \centering
  \begin{subfigure}{0.49\linewidth}
    \includegraphics[width=1.0\linewidth]{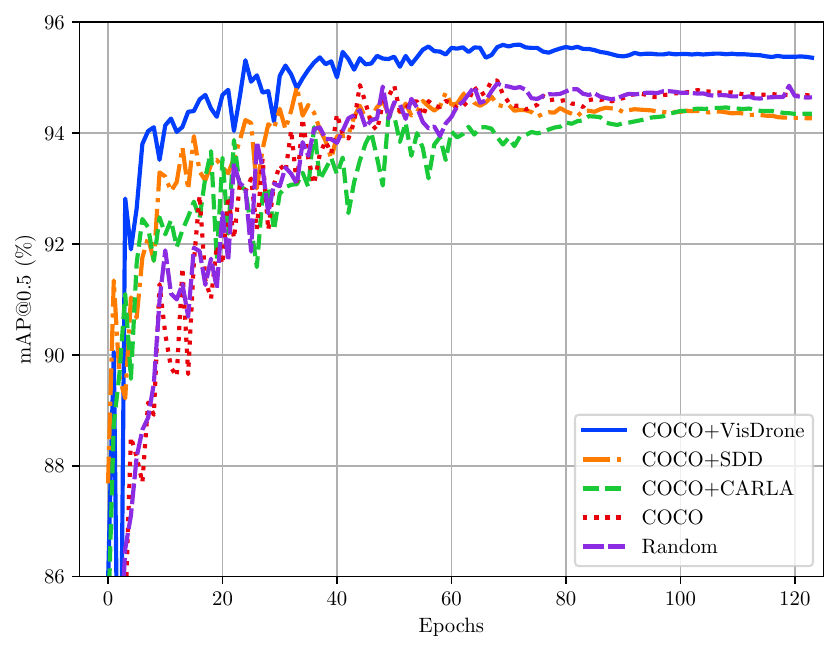}
    \caption{YOLOv8x Models}
    \label{fig:comp-a}
  \end{subfigure}
  \hfill
  \begin{subfigure}{0.49\linewidth}
    \includegraphics[width=1.0\linewidth]{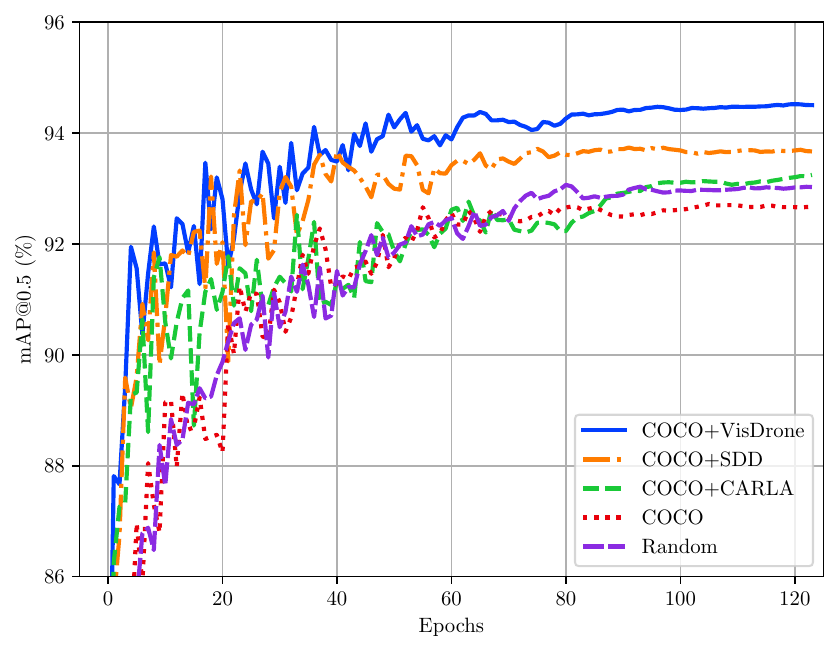}
    \caption{YOLOv8n Models}
    \label{fig:comp-b}
  \end{subfigure}
  \caption{Comparison of the model performance for (a) YOLOv8x, and (b) YOLOv8n (b) models for different pretraining schemes. Performance for the two model architectures follows similar trends with VisDrone pretraining achieving better results for both architectures.}
  \label{fig:xn_comparison}
\end{figure*}

%% file: sec/3_dataset.tex
\section{Dataset Creation and Analysis}

Constellation is a dataset of images of a typical street intersection in an urban metropolis, acquired by a high-altitude camera installed on a side of a building. The dataset contains images collected in two different periods: years 2019/2020 and 2023. There are a total of 28 different time intervals (in length between 1 minute and 1 day) in the dataset, with different times of day, time of year, weather, and background qualities. 

The dataset contains $13,314$ human-annotated images with bounding box annotations for pedestrians and vehicles. An overview of the dataset grouped by weather condition, with the suggested training/test set splits we use for evaluation is provided in Table \ref{tab:weather_splits} with the number of images per split. 
The dataset is split into 5 conditions: (i) overcast, (ii) sunny, (iii) night, (iv) sunny with sharp shadows, and (v) foggy. The foggy images in the dataset correspond to a rare day with wildfire smoke. %

\begin{table}[]
\caption{Training and testing dataset sizes, grouped by the different weather conditions.}
\centering
\begin{tabular}{|l|l|l|}
\hline
\textbf{Weather}  & \textbf{Split} & \textbf{\# of Images} \\ \hline
Overcast          & Train          & 1,205            \\ \cline{2-3} 
                  & Test           & 602             \\ \hline
Sunny               & Train          & 7,355            \\ \cline{2-3} 
                  & Test           & 1,282            \\ \hline
Night             & Train          & 1,600             \\ \cline{2-3} 
                  & Test           & 450             \\ \hline
Sun/Sharp Shadows & Train          & 70              \\ \cline{2-3} 
                  & Test           & 150             \\ \hline
Foggy             & Train          & -               \\ \cline{2-3} 
                  & Test           & 600             \\ \hline
\end{tabular}
\label{tab:weather_splits}
\end{table}

\paragraph{Dataset Description}

An overview of the dataset is shown in Figure \ref{fig:constellation_overview}. The dataset contains data collected at all times of day (a-d), which results in a large variability in lighting conditions as sharp shadows are cast by buildings over the intersection depending on the time of day. These shadows result in large brightness changes that object detection models need to take into account (for example scenes shown in (a) and (g)). The background is highly variable due to the changes in shadows and lighting, frequent changes to the scene background, birds flying over the intersection, parked cars, and waiting pedestrians. Moreover, the foreground is difficult to estimate given the small size of the shadows and their similarity to pedestrians wearing dark colors. Frames cover all seasons of the year, and weather conditions that include (i) sunny, (ii) overcast, and (iii) rain.

The dataset comprises of $13,314$ annotated frames, divided into $10,230$ training and $3,084$ test images collected at $28$ different time intervals. The raw frames are captured from the side of a building with a viewing angle that is slightly less than perpendicular to the street plane. 
The frames are transformed using a perspective transformation such that the viewing angle to the street is exactly $90$ degrees, to represent the bird's-eye view perspective. 
The images in the dataset have a resolution of 832x832, with the intersection centered in the frame. The original images are captured in 1920x1080 resolution and are subsequently transformed to obtain a top-down view and cropped to square images. Following the input size scaling rules for YOLO models and to enable low processing times without relying on sliced inference techniques~\cite{akyon2022slicing}, the input image size is chosen as 832x832. For all frames, axis-aligned bounding boxes are provided for pedestrian and vehicle classes. All annotations were done in two rounds where the work of an initial annotator was reviewed by a second annotator.

Eighteen of the time intervals are obtained from videos annotated at 10 frames per second (FPS) in years 2019\&2020. Five of these contain 300 frames, and the other 13 contain 450 frames - for a total of 7,350 annotated frames. In addition to these, the dataset contains 5,964 frames captured and annotated at 20\thinspace{s} intervals in 10 different scenes during 2023 to increase variability in the subject, lighting, weather, time-of-year, and background conditions.

Constellation features many challenging frames with large crowds where picking individuals apart is difficult; we show some examples of such scenes in Figure \ref{fig:high_overlap}, where annotation requires thinking and care by human annotators.

\textbf{Training Set} The training set consists of 10,230 frames divided into 19 subsets collected on different days with different weather and time-of-day conditions. 15 of the subsets are from 2019 and 2020 (corresponding to 6,160 frames), and 4 are from 2023 (corresponding to 4,070 frames). All frames are collected from an intersection in Manhattan.  %

\textbf{Test Set} The test set consists of 3,084 frames divided into 9subsets collected on different days with different weather and time-of-day conditions. This test set contains 12,399 vehicles and 14,229 pedestrian bounding boxes. 3 of the subsets are from years 2019 and 2020 (corresponding to 1,201 frames), and 6 are from 2023. 5 of the subsets from 2023 feature the intersection before repavement. Each subset contains 70 to 2,000 frames. As part of the dataset, we provide splits of data depending on the acquisition year for studies involving the transfer of knowledge from old frames to new frames.

\textbf{Pseudo-labeled Data Collection} After the initial training of object detection models based on the data above, we used the YOLOv8x model in inference mode and collected 50,557 more images following a tracking-by-detection paradigm, using the tracked objects to provide additional annotations \cite{lee2013pseudo}. To deal with scenarios where an object is not detected, we mask the image to keep only the pixels where an object was detected (with 8-pixel padding) and use the median of the past 120 seconds (captured at 1FPS) as the background pixels for the rest of the image. For generating this dataset, during data collection, we use the ByteTrack tracking algorithm~\cite{zhang2022bytetrack}.

%% file: sec/4_experiments.tex
\section{Experiments}

We explored the performance of object detection models in terms of accuracy in detection and
inference time on our dataset over five experiments: (i) model architecture, (ii) pretraining dataset and image augmentation, (iii) model scaling, (iv) semi-supervised training, and (v) performance drift. For computation, we use eight A100 NVIDIA GPUs with 40 GB VRAM and a batch size of 16. For YOLOv8x models trained at 2x resolution, we use a batch size of 8 during training due to the higher memory requirements. Since we focus on small object detection and the sensitivity of IoU to small bounding boxes, we report AP and mAP for  Intersection over Union (IoU) value of of 0.6.

\begin{table*}[h!]
\caption{Per-class average precision, mean average precision, and inference time for each model for different architectures, using models pretrained on the COCO dataset.}
\centering
\begin{tabular}{|l|l|l|l|l|} 
\hline
\textbf{Model Name}                     & \textbf{Pedestrian AP@0.5} & \textbf{Vehicle AP@0.5} & \textbf{mAP@0.5} & \textbf{Inference Time (ms)}  \\ 
\hline
YOLOv8x \cite{yolov8_ultralytics}                                & 87.4                       & 98.6                    & 93.0             & 11.5                          \\ 
\hline
YOLOv8n (2x)                            & 91.2                       & 98.5                    & 94.8             & 7.2                           \\ 
\hline
YOLOv8x (2x)                            & 91.2                       & 98.4                    & 94.8             & 43.6                          \\ 
\hline
YOLOv8n (2x+SR)                         & 90.1                       & 98.4                    & 94.2             & 7.2                           \\ 
\hline
YOLOv8x (2x+SR)                         & \textbf{91.3}              & \textbf{98.7}           & \textbf{95.0}    & 43.6                          \\ 
\hline
YOLOv8x (P2)                            & 89.5                       & 98.6                    & 94.0             & 15.1                          \\ 
\hline
YOLOv8x (P6)                            & 89.4                       & \textbf{98.7}           & 94.0             & 7.6                           \\ 
\hline
YOLOv8x (P2-P6)                         & 89.9                       & \textbf{98.7}           & 94.3             & 24.5                          \\ 
\hline
DETR-l \cite{lv2023detrs}                                 & 86.5                       & 98.1                    & 92.6             & 9.8                           \\ 
\hline
DETR-x \cite{lv2023detrs}                                 & 87.3                       & 97.8                    & 92.3             & 14.5                          \\ 
\hline
YOLOv7 \cite{wang2023yolov7}                                  & 78.8                       & 98.2                    & 92.7             & 15.4                            \\ 
\hline
CFINet \cite{yuan2023small}                                  & 82.8                       & 95.8                    & 89.3             & 31.4                          \\ 
\hline
\textcolor[rgb]{0.114,0.11,0.114}{STD \cite{yu2023spatial}} & 79.9                       & 90.8                    & 85.3             & 14.2                          \\ 
\hline
LSKNet-S \cite{li2023large} & 64.8                       & 90.2                    & 77.5             & 31.1                           \\ 
\hline
PPYOLOE+ \cite{xu2022pp}                                & 88.5                       & 97.7                    & 93.1             & 21.6                          \\ 
\hline
RTMDet \cite{lyu2022rtmdet}                                  & 87.5                       & 98.3                    & 92.9             & 24.5                          \\
\hline
\end{tabular}
\label{tab:results_architecture}
\end{table*}

\begin{table}[]
\caption{Comparison between the default augmentations in YOLOv8 and the new set of augmentations proposed for Constellation.}
\scalebox{0.95}{
\begin{tabular}{|c|c|}
\hline
\textbf{YOLOv8} & \textbf{Constellation} \\ \hline
Blur                          & Blur                                 \\ \hline
MedianBlur                    & -                                    \\ \hline
ToGray                        & ToGray                               \\ \hline
CLAHE                         & -                                    \\ \hline
RandomBrightnessContrast      & RandomBrightnessContrast             \\ \hline
RandomGamma                   & -                                    \\ \hline
ImageCompression              & -                                    \\ \hline
                              & MotionBlur                           \\ \hline
                              & RandomShadow                         \\ \hline
                              & RandomRain                           \\ \hline
\end{tabular}}
\label{tbl:augmentations}
\end{table}

\begin{table*}[h!]
\centering
\caption{Per-class average precision and mean average precision for different pretraining dataset choices for the YOLOv8x architecture.}
\begin{tabular}{|l|l|l|l|}
\hline
\textbf{Pretraining Type (Without Augmentation)} & \textbf{Pedestrian AP@0.5} & \textbf{Vehicle AP@0.5} & \textbf{mAP@0.5} \\ \hline
COCO                                             & 87.4                       & \textbf{98.6}           & 93.0             \\ \hline
COCO+CARLA                                       & 88.6                       & 98.2                    & 93.4             \\ \hline
COCO+VisDrone                                    & \textbf{90.9}              & \textbf{98.7}           & \textbf{94.8}    \\ \hline
COCO+SDD                                         & 87.4                       & \textbf{98.6}           & 93.0             \\ \hline
Random Initialization                            & 92.5                       & 98.3                    & 92.5             \\ \hline
\end{tabular}
\label{tab:results_pretraining}
\end{table*}

\begin{table*}[h!]
\centering
\caption{Effect of augmentation on pretraining with a YOLOv8x model after 125 epochs. Mean average precision and per-class average precision are shown.}
\begin{tabular}{|l|l|l|l|}
\hline
\textbf{Pretraining Type (With Augmentation)} & \textbf{Pedestrian AP@0.5} & \textbf{Vehicle AP@0.5} & \textbf{mAP@0.5} \\ \hline
COCO                                          & 90.7                       & 98.6                    & 94.7             \\ \hline
COCO+CARLA                                    & 90.1                       & 98.7                    & 94.4             \\ \hline
COCO+VisDrone                                 & \textbf{92.0}              & \textbf{98.8}           & \textbf{95.4}    \\ \hline
COCO+SDD                                      & 90.0                       & 98.6                    & 94.3             \\ \hline
Random Initialization                         & 90.6                       & 98.7                    & 94.7             \\ \hline
\end{tabular}
\label{tab:results_pretraining_aug}
\end{table*}

\begin{table*}[h!]
\centering
\caption{Per-class average precision and mean average precision for the final training with the pseudo-labeled data (BoxMask) used for pretraining.}
\begin{tabular}{|l|l|l|l|}
\hline
\textbf{Pretraining Type}      & \textbf{Pedestrian AP@0.5} & \textbf{Vehicle AP@0.5} & \textbf{mAP@0.5} \\ \hline
BoxMask+Constellation & 90.1           & 98.4                 & 94.3         \\ \hline
\end{tabular}
\label{tab:results_architecture_automated}
\end{table*}

\begin{table*}[h!]

\caption{Per-class average precision and mean average precision for two models trained using the two different subsets of the training set: using (i) only data from 2019/2020, and (ii) only data from 2023, and similarly analyze their performance on the test set.}
\centering
\begin{tabular}{|l|l|l|l|l|}
\hline
\textbf{Training Dataset} & \textbf{Test Split} & \textbf{Pedestrian AP@0.5} & \textbf{Vehicle AP@0.5} & \textbf{mAP@0.5} \\ \hline
                          & All                 & 76.7                       & 95.8                    & 86.3             \\ \cline{2-5} 
2019/2020                 & 2019/2020           & \textbf{74.4}              & \textbf{99.0}           & \textbf{86.7}    \\ \cline{2-5} 
                          & 2023                & 77.8                       & 93.5                    & 85.6             \\ \hline
                          & All                 & \textbf{82.3}              & \textbf{97.2}           & \textbf{89.7}    \\ \cline{2-5} 
2023                      & 2019/2020           & 59.7                       & 97.2                    & 78.5             \\ \cline{2-5} 
                          & 2023                & \textbf{88.1}              & \textbf{97.5}           & \textbf{92.8}    \\ \hline
\end{tabular}
\label{tab:results_drift}
\end{table*}

\subsection{Evaluating Model Architectures}

 Different models offer different trade-offs between detection accuracy and inference time. For object detection experiments, we explored YOLOv8, RT-DETR, and CFINet models~\cite{wang2022yolov7,yolov8_ultralytics,lv2023detrs}. An alternative to the one-stage object detectors considered, CFINet integrates Faster R-CNN with a coarse-to-fine region proposal network and feature imitation learning~\cite{yuan2023small}. As an alternative to models that were trained at the input resolution (832x832), we also explore models trained at 2x resolution. To do so, we naively upscale our input images to 1664x1664 resolution using bilinear interpolation or by using Real-ESRGAN \cite{wang2021real}. This allows for the CNN layers to process small objects with better precision while avoiding the large inference time cost of applying super-resolution. Table \ref{tab:results_architecture} shows the results of trying these different architectures. All models are trained for 125 epochs. 2x refers to models trained using bilinear interpolation, while 2x+SR refers to models trained using Real-ESRGAN. As the use of Real-ESRGAN adds a sizable inference time latency, we give results for two separate models. P2, P6 and P2-P6 refer to YOLO architectures using different detection layers; default models use layers P3-P5, P2 models use layers P2-P5, P6 models use layers P3-P6. We propose a modified architecture that uses layers P2-P6 \cite{lin2017feature,yolov8_ultralytics}.

 Although we achieve the best results when training the models with high-resolution inputs, we see competitive results at real-time constraints with the modified YOLOv8x architectures.

\subsection{Evaluating Pretraining Datasets} 

Pedestrians and vehicles are classes with high amounts of variance. To help generalize our models better to novel situations, we explored the integration of Constellation with previously released datasets in literature to improve model performance. To this end, we considered the drone-view VisDrone and Stanford Drone datasets and simulated 3D data generated using the CARLA simulator. 
We give an overview of the datasets we consider for pretraining in Figure \ref{fig:different_datasets}. 

\textbf{VisDrone} The VisDrone object detection challenge dataset contains 7,019 images~\cite{zhu2018visdrone}. VisDrone images are a mix of top-down and ground-level data with different perspectives and zoom levels. The dataset contains many different object classes, which we convert to our 2-class setup. The dataset contains 175,551 vehicles and 88,261 pedestrians.

\textbf{Stanford Drone Dataset} Stanford Drone Dataset consists of 60 videos collected from 8 different scenes~\cite{robicquet2016learning}. We use 15 of the videos with a high pedestrian, biker, and vehicle concentration. By moving a sliding window with 20\% overlap, we convert each input frame into a set of 832x832 frames. Post-processing, the dataset has 42,727 images containing 222,456 pedestrians and 35,587 vehicles.

\textbf{CARLA} We use the CARLA simulator to generate additional synthetic frames with ground truth annotations~\cite{dosovitskiy2017carla}. We replicated the placement of the camera artificially in CARLA and used the same perspective transformation as shown in Figure \ref{fig:different_datasets}(b), 
and collected 26,984 frames with the same perspective transformation. The dataset contains 91,848 pedestrians and 67,815 vehicles.

\textbf{Results} For pretraining experiments, we initialize and train YOLOv8x models for 125 epochs on the chosen pretraining dataset: (i) VisDrone, (ii) SDD, and (iii) CARLA. After obtaining these pretrained models, we fine-tuned the model on Constellation for 125 epochs. 

Experiments for different pretraining schemes are presented in Table \ref{tab:results_pretraining}. Due to the similar performance among all pretraining schemes, we choose to use the COCO-trained models. Surprisingly, the CARLA-generated dataset does not obtain competitive performance despite the compatibility of the setup for the overlapping pedestrian detection problem, indicating a big gap between simulation and real-world performance for this specific problem \cite{kadian2020sim2real}. Comparing the visual features between CARLA renders and the real-world intersection, we believe this is due to the inadequate realism of CARLA renders when used for training.

The detection results are promising. However, for the application scenarios of high-altitude pedestrian and vehicle detection where a single detection error could be catastrophic (for example, safety warnings and navigation for disabled individuals), the AP values of the baseline models for pedestrians are insufficient for real-world deployment.

\subsection{Image Augmentation} 
We assumed that scenes with varying background conditions would benefit from augmentation approaches which are different compared to classical object detection augmentations, which have to consider varying image quality and choose augmentations correspondingly. We explored custom image augmentation approaches tuned for the use case to improve model robustness on other out-of-distribution images. 

The default set of augmentations for YOLOv8 and the set of augmentations that we propose are listed in Table \ref{tbl:augmentations}. %
We use the Albumentations library to add the new augmentations~\cite{2018arXiv180906839B}. RandomShadow creates random transparent patches of darkness in the input simulating shadow, which we postulate should help the model anticipate the sudden changes in brightness levels due to crisp shadows cast by buildings. MotionBlur adds random motion blur and helps simulate the camera blur for fast-moving entities or the possible effects of raindrops on the camera lens. Random rain adds an artificial rain effect on top of the scene and is intended to make the model robust to frame-specific corruption during such weather. 
We use these augmentations with the models trained for the different pretraining datasets to explore whether the addition of these augmentations could improve model performance.

Table \ref{tab:results_pretraining} shows the results of various models on the Constellation dataset and Table \ref{tab:results_pretraining_aug} shows the effect of adding augmentations. We find a consistent minor improvement in AP scores for all models with the added augmentations, which suggests that simulation of different weather conditions might enable the building of more accurate street-level object detectors. Our best-performing models use the VisDrone dataset for pretraining, resulting in 4.6\% improvement in pedestrian AP. Importantly, for vehicles that are larger objects, the pretraining has very little effect, corresponding to only a improvement of 0.1\% AP over the COCO-pretrained model. CARLA and SDD datasets underperform compared to VisDrone as pretraining datasets.

 \subsubsection{Model Scaling}

 To examine the trends in model scaling, we plot mAP@0.5 per epoch for training with YOLOv8x vs. YOLOv8n models in Figure \ref{fig:xn_comparison}. We observe the same trends and consistent relative ordering of the performance curves in both plots, implying that model comparison can be conducted with smaller, faster-to-train networks. Moreover, we find only a small gap in mAP between the two model architectures for the best-performing models, corresponding to a difference in mAP of 1.1\%. As intersection monitoring will benefit from edge computing devices deployed next to cameras to reduce bandwidth and preserve anonymity, this small difference shows the suitability of object detection models in edge deployments. 

 \subsubsection{Semi-Supervised Object Detection}

 To improve the performance of the models, we deployed the YOLOv8x model trained on Constellation in inference mode for real-time data tracking and bounding box collection as an augmentation approach. %
 Using the tracking-by-detection paradigm allowed the model predictions to avoid false positives and false negatives \cite{bewley2016simple}. Another significant challenge when using such a model for automatic data collection is dealing with objects that have not been detected. To reduce this false negative rate in data collected in this manner, we masked all areas of the image beyond the bounding box bounds with a median image calculated over the last 90 seconds of the video stream.

 The unlabeled data for semi-supervision was collected over 4 days, with one frame being saved every 20 seconds. We refer to this dataset as BoxMask. We use this dataset for a 125-epoch pretraining step, followed by 125 epochs of training on human-annotated Constellation images. The results are presented in Table \ref{tab:results_architecture_automated}. We find that this dataset achieves a similar performance to VisDrone, suggesting a pretraining stage with more diverse pseudo-labeled data could be used as a replacement for standard pretraining datasets. 
 
\subsection{Performance Drift}

We explored the effect of removing portions of the training data depending on the year of acquisition from the training set, to look at how changes in street-level conditions affect models over time. We ran two experiments (i) using the data from 2019/2020 only, (ii) using the data from 2023 only. The same split is used for the test set. YOLOv8x models are trained for 125 epochs on both splits and test each model on both splits. We summarize the experiments in Table \ref{tab:results_drift}. The models trained with split (ii) perform significantly better in the corresponding test set split.
We thus postulate that for deploying object detection models, regularly updated data collection and refinement need to be a standard procedure. Future research could focus on minimizing the performance difference between splits (i) and (ii) through novel training or augmentation strategies.

%% file: sec/5_ethics_ack_conclusion.tex
\section{Ethical Considerations} 

The Constellation dataset is collected using a camera deliberately installed at an altitude that makes it impossible to discern the pedestrian faces or to read car license plates. The academic institution at whose facilities the camera is installed provided an IRB waiver for sharing this ``high elevation data" with the general public since the data inherently preserves privacy.

\section{Conclusion}

We created and described the Constellation dataset which consists of $13K$ images of traffic intersection scenes acquired by a high-elevation camera deployed in a metropolis.
The purpose of the dataset is to support research on high-altitude object detection in dense urban environments, which is a key component for AI-based smart city applications. 
The analysis of the performance of object detection models using this dataset highlighted the difficulties in detecting small pedestrians in traffic intersections with changing backgrounds, lighting, building shadows, and weather conditions.
The experiments with pretraining on similar datasets show that strategic data augmentation combined with pretraining can improve detection accuracy. The best models yield promising results, with the top-performing model achieving a pedestrian detection AP of 92.0\% and an overall mean average precision (mAP) of 95.4\%. We note that there remains a large gap between pedestrian and vehicle detection performance, which needs to be closed for the deployment of safety-critical intersection monitoring models for analytics and tracking applications.
We anticipate that Constellation will help advance object detection methods for high-altitude pedestrian and vehicle monitoring for applications such as safety warning generation and collection of traffic analytics. By releasing the dataset along with the trained baseline models and the corresponding codebase we aim to facilitate future research and applications in urban safety applications.

\section*{Dataset Availability} 

We make the code and datasets available at \url{https://github.com/zk2172-columbia/constellation-dataset}.

\section*{Acknowledgements}

This work was supported in part by NSF grants CNS‐1827923, OAC‐2029295, CNS-2148128, EEC-2133516, NSF grant CNS‐2038984 and corresponding support from the Federal Highway Administration (FHA), ARO grants W911NF2210031 and W911NF1910379.